\documentclass{article}

\usepackage{iclr2027_conference,times}

\usepackage{amsmath,amsfonts,bm}

\def\eqref#1{equation~\ref{#1}}

\def\1{\bm{1}}

\DeclareMathAlphabet{\mathsfit}{\encodingdefault}{\sfdefault}{m}{sl}
\SetMathAlphabet{\mathsfit}{bold}{\encodingdefault}{\sfdefault}{bx}{n}

\usepackage{booktabs}
\usepackage{graphicx}
\usepackage{hyperref}
\usepackage{microtype}
\usepackage{url}
\usepackage{xcolor}
  \usepackage{wrapfig}
\usepackage{needspace}
\title{\benchmark: Benchmarking Spatial Agentic Intelligence in the Wild}

\newcommand{\projectlink}[2]{\begingroup\edef\linktarget{#1}\ifx\linktarget\empty #2\else\href{#1}{#2}\fi\endgroup}
\author{Yuxuan Cao$^{*}$\quad Junlong Li$^{*}$\quad Hao Li\quad Junxian He\\
The Hong Kong University of Science and Technology\\
\normalfont\small \texttt{\{ycaoce, jlini\}@connect.ust.hk}\\
\normalfont\small Project Page: \url{https://mine-odyssey.github.io/}\\
\normalfont\small Code: \url{https://github.com/hkust-nlp/MineOdyssey}}

\newcommand{\benchmark}{\textsc{Mine Odyssey}}
\newcommand{\nummaps}{30}
\newcommand{\numoutdoormaps}{20}
\newcommand{\numindoormaps}{10}
\newcommand{\numtasks}{180}

\newcommand{\numwaypoints}{1,059}
\newcommand{\numtaskwaypoints}{715}

\iclrfinalcopy

\begin{document}

\maketitle
\lhead{}
\begingroup
\renewcommand{\thefootnote}{\fnsymbol{footnote}}
\footnotetext[1]{Equal contribution.}
\endgroup

\begin{abstract}
Advances in foundation models are driving efforts to introduce agents to  assist people in the physical world. Such agents require agentic spatial intelligence: 
exploring unfamiliar environments, updating spatial understanding through interaction, and adapting actions based on feedback to sustain progress toward a sequence of goals.
{Evaluating these capabilities requires environments that reflect the diversity and structural complexity of real-world places, yet existing benchmarks typically cover only a limited range of spatial layouts, scales, and traversal requirements.
To address this gap, we introduce \textbf{\benchmark{}}, a benchmark for evaluating agentic spatial intelligence using Minecraft reconstructions of real-world locations. It comprises \numtasks{} tasks covering 30 such locations across 20 countries and regions on five continents, including 20 outdoor and 10 indoor settings. These settings are selected to provide broad coverage across spatial scales, layouts, terrains, and connectivity patterns, ranging from the urban streets of Midtown Manhattan to the rural Entrup, and from the steep terrain of Santa Lucía Hill to Buckingham Palace. Within each reconstruction, we select meaningful locations as waypoints (\emph{e.g.}, historic landmarks, buildings, and rooms) and manually verify their accessibility. Each task provides a natural-language instruction specifying which waypoints to visit and in what order. Completing these tasks requires agents to find accessible routes and entrances, open doors, and move between levels using stairs and ladders, while monitoring their progress and recovering from navigation errors.}
 Across eight evaluated state-of-the-art models, GPT-6 Astra achieves the highest success
  rate of 85.6\%.
  However, the second-best model, Claude Opus 5.5, completes 73.9\% of tasks, while the strongest evaluated open-weight model, DeepSeek-V4.1-Flash, reaches 23.9\%, highlighting substantial room for improvement in the agentic spatial intelligence of current models. 
Comprehensive analyses and ablation studies on \benchmark{} reveal current models’ limitations and provide insights for advancing agentic spatial intelligence.
\end{abstract}

\section{Introduction}
Advances in foundation models are opening new possibilities
for agents to move beyond digital workflows and interact with the
physical world~\citep{geminiroboticsteam2025gemini}.
Recent evaluations of frontier models such as GPT-6 Astra \citep{openai2026gpt6astra}
using robot-control tools have begun to explore this possibility in
embodied settings~\citep{robodojo2026gpt6astra}.
Potential applications for such agents range from robotic guide dogs
that assist people with visual impairments~\citep{hayamizu2026woofs}
to autonomous delivery robots that transport
goods~\citep{wang2025openbench}.
In these settings, agents must adapt their movements and interactions
based on the spatial relationships in their surroundings, including
how spaces connect and which objects and locations are reachable.
We refer to this capability as \emph{agentic spatial intelligence}:
the ability to proactively acquire and continually update spatial
information through interaction, use it to guide observation and
action, and adapt those actions as new evidence
emerges~\citep{pei2026sapnav,deng2026spacevln}.

Existing benchmarks often evaluate agents in relatively narrow settings, such as household environments
~\citep{yang2025embodiedbench,qiao2025navbench}
or urban navigation
~\citep{wang2025openbench,dalal2026citynav,wang2026cityseeker,ju2026urbanground}, leaving their ability to operate across a broader range of spatial structures and traversal requirements insufficiently tested.  However, real-world tasks often arise in environments with varied layouts and traversal requirements.
An agent may need to follow irregular paths through a rural
  settlement, find a traversable route across a steep hillside,
  or navigate entrances, corridors, and multiple levels within
  a palace or stadium.
  These environments differ not only in appearance, but also
  in scale, connectivity, access constraints, and elevation.
  This diversity, together with the need to maintain spatial
  information and revise plans throughout execution, makes
  reliable performance across environments particularly challenging.
\begin{figure}[t]
\centering
\includegraphics[width=\linewidth]{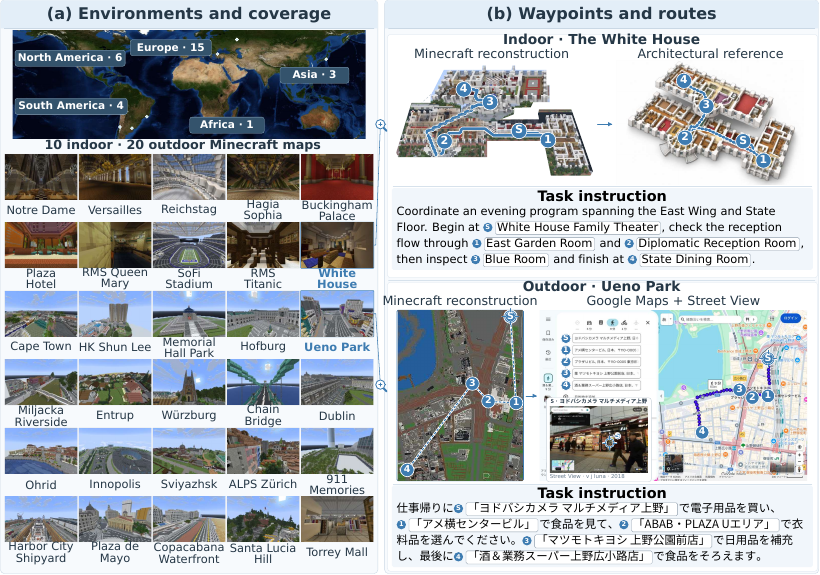}
\caption{Overview of \benchmark{} environments and task construction.
(a) \nummaps{} Minecraft reconstructions and geographic coverage.
(b) Task construction with
 White House and Ueno Park examples: reference locations are mapped to waypoints and ordered into navigation tasks.
S denotes the start; numbered markers and dashed lines indicate visit order. 
 }    \vspace{-\baselineskip} 
\label{fig:task-construction-overview}
\end{figure}

  To  address these challenges, we introduce \benchmark{},  a benchmark
  for evaluating agentic spatial intelligence through diverse, realistic, and long-horizon navigation tasks in Minecraft reconstructions of real-world locations. {Here we choose Minecraft as our evaluation platform for its rich ecosystem of community-created mods and world saves. Detailed reconstructions of cities, landscapes, and buildings from around the world provide diverse spatial environments. The platform also allows episodes to be reset to specified initial conditions and task completion to be verified automatically from the agent's in-game position.}
 \benchmark{} spans \nummaps{} real-world locations recreated
  in Minecraft, with \numtasks{} tasks across 20 countries and
  regions on five continents, including \numoutdoormaps{} outdoor and
  \numindoormaps{} indoor environments.
  To capture the diversity and structural complexity of real-world spatial environments, we curate spots spanning dense urban districts, rural settlements, mountainous terrain, palace complexes, stadiums, and historic passenger ships. Representative examples include Midtown Manhattan, Cape Town, Entrup, Santa Lucía Hill, Buckingham Palace, SoFi Stadium, and RMS Queen Mary.
  These places provide variation in street connectivity, terrain, room
  arrangement, entrances, and vertical circulation.
 Minecraft makes these structures interactive through first-person movement,
  collision, and actions such as opening doors and climbing ladders and stairs.
  
Since these maps don't come with annotated destinations for evaluation, we mark named locations as waypoints that serve as destinations or intermediate stops in navigation tasks. To preserve realism and semantic relevance, we select recognizable places such as historic landmarks, palace rooms, stadium facilities, and building entrances rather than arbitrary points on the map. Each task is then defined as an ordered sequence of these waypoints, requiring agents to find feasible paths and execute the movements and interactions needed to reach each destination in order.

We also provide a programmable agent framework that connects
  visual observations with game APIs and keyboard--mouse
  control, allowing agents to compose and execute their own
  action sequences. Agents can inspect intermediate feedback
  and interrupt ongoing commands, supporting closed-loop
  interaction as the environment evolves. Isolated execution
  prevents access to hidden task data and interference between
  runs, while independent arrival checks support automated,
  parallel evaluation.

Our evaluation reveals substantial differences in task success and
decision efficiency. GPT-6 Astra \citep{openai2026gpt6astra} achieves an 85.6\% task success rate,
compared with 73.9\% for Claude Opus 5.5~\citep{anthropic2026claudeopus55} and 23.9\% for the strongest
open-weight model, DeepSeek-V4.1-Flash \citep{deepseek2026v41flash}.
Performance differences also depend on the environment, with a larger
success-rate gap between Astra and Opus 5.5 indoors than outdoors.
In more than half of failed attempts, agents do not reach even the first
required destination. We further conduct ablation analysis and case studies to better understand  the strengths and limitations of current agents.
Our task suite and evaluation framework will also be publicly released
to support future research on agentic spatial intellgence in diverse
real-world environments. 
\section{Tasks Environment and Evaluation Framework}
\subsection{Environments}
\paragraph{Game Details}\label{sec:game-details}
Minecraft has been widely adopted as a research platform with its interactive 3D environments and consistent physical mechanics~\citep{guss2019minerl,fan2022minedojo,guo2025mineworld,ju2026mineexplorer}.
Its open community ecosystem also allows us to integrate diverse, community-built worlds into a unified testbed. 
Diverse mods provide fine-grained control over agent observation and interaction and enable automatic verification of task outcomes using in-game agent positions.
Our experiments adopt \href{https://www.minecraft.net/en-us/article/minecraft-java-edition-1-21-11}{ Minecraft Java Edition
1.21.11}
with  \href{https://neoforged.net/news/21.11release/}{NeoForge 21.11.44}. We also install these mods:
  \href{https://modrinth.com/mod/xaeros-world-map}{\emph{{Xaero's World Map}} 1.44.2} for full-screen map,
  \href{https://modrinth.com/mod/xaeros-minimap}{\emph{Xaero's Minimap} 26.4.2} for waypoint markers,
  \href{https://modrinth.com/mod/bocchud}{\emph{BoccHUD} 0.4.1} with \href{https://modrinth.com/mod/mafglib}{\emph{MaFgLib} 0.4} for HUD, and
  \href{https://modrinth.com/mod/lambdynamiclights}{\emph{LambDynamicLights} 4.9.1} for dynamic lighting.
  By default, agent in each task begins at a specified position and orientation
  in first-person view, with a nominal $100^\circ$ field of view,
  frozen noon lighting, clear weather, Adventure mode, Peaceful
  difficulty, and only the \emph{Xaero's World Map} and \emph{LambDynamicLights} are enabled.
  Additional configurations and restrictions are detailed in
  Appendix~\ref{app:settings}.
\begin{figure}[t]
\centering
\includegraphics[width=\linewidth]{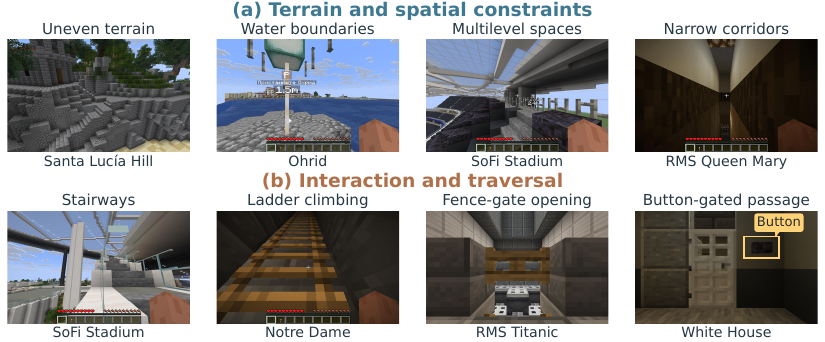}
\caption{ Representative terrain, spatial constraints, and interactions across the diverse collection of maps in MINE ODYSSEY.}
\label{fig:terrain-interactions}
\end{figure}

\paragraph{Maps }
                            
We source our maps from BuildTheEarth,\footnote{\url{https://buildtheearth.net}} a global
community project dedicated to reconstructing real-world locations in
Minecraft. 
We aim to evaluate agents in environments that reflect the diversity and
spatial realism of real-world places while preserving experimental control.
As shown in Figure~\ref{fig:task-construction-overview}, we use \nummaps{} existing Minecraft builds modeled on locations
across five continents, spanning 20 different countries and regions. More detailed statistics and can be found in Appendix~\ref{app:map-collection}.
The collection includes \numoutdoormaps{} outdoor and \numindoormaps{} indoor
environments. Outdoor maps cover residential and commercial districts, parks,
hillsides, and waterfronts, varying in street connectivity, landmark density,
and relief.
Indoor maps include palaces, religious buildings, hotels, stadiums, and
passenger ships, with variation in room arrangement, corridor structure,
entrances, and vertical circulation. 
These environments present diverse terrain and spatial constraints,
including uneven ground, water boundaries, narrow corridors, and multilevel
spaces. Navigating them can involve ascending stairs, climbing ladders,
opening fence gates, or activating buttons to pass through doors, as
illustrated in Figure~\ref{fig:terrain-interactions}.
Versioned world saves allow each task to begin from a consistent map state.

\paragraph{Waypoints}
We use waypoints to represent the destinations and intermediate stops that an
agent may be asked to visit. Each waypoint pairs a human-readable place name
with a three-dimensional arrival position in a specific Minecraft world. Our
waypoints include landmarks, shops, public facilities, transport entrances,
and indoor rooms, covering destinations with different functions and access
conditions.
We construct each waypoint in three stages. First, an annotator identifies a
named place and its approximate real-world location using geographic or
venue-specific references, including Google Maps,
OpenStreetMap, and site maps. Second, the annotator finds the corresponding
place in the Minecraft reconstruction by matching the surrounding street or
room layout and visible cues such as building facades, distinctive structures,
and signs. Third, the annotator enters the world, visits the matched place, and
creates the waypoint at an observed, reachable standing position on the
correct floor. 
During task execution, the agent receives the waypoint name and can access its visual map
marker via map, whereas the precise three-dimensional arrival position remains evaluator-only. 
{ A waypoint marker will appears in the first-person view when the agent is within 32 blocks of the waypoint, providing local visual guidance toward the destination.}
The catalog contains \numwaypoints{} waypoints, of which
\numtaskwaypoints{} distinct identifiers occur in the task pool. 

\paragraph{Tasks}
As shown in Figure~\ref{fig:task-construction-overview}(b), each task specifies a map, a starting waypoint and orientation, and an ordered sequence of destinations described in a natural-language instruction. The agent must identify and visit these destinations in order, choosing its own route using the permitted controls and ending at the final waypoint.
To construct tasks, we provide large language models with waypoint names, place types, and map descriptions to generate candidate task instructions and stop orders with  a specific role and purpose. 
We then manually select and
refine these candidates based on three criteria:
\textit{semantic coherence}, so that the stops and their order fit the stated
purpose; \textit{route plausibility}, avoiding redundant visits, unnecessary
backtracking, and implausibly short segments; and \textit{diversity}, covering
varied roles, scenarios, and navigation requirements. To ensure task
diversity, we vary starting points, distances between stops, elevation changes,
and interactions such as opening doors, pressing buttons, or climbing ladders,
while limiting repeated ordered waypoint pairs within each map.
\ yx {To guarantee task feasibility, we first use Baritone (a Minecraft pathfinding and navigation mod) for a preliminary reachability check, then manually traverse the task routes in the game to verify
the reachability.}
The resulting collection contains \numtasks{} tasks with
2--12 stops each.
\subsection{Verifier.}
\label{sec:evaluation-protocol}
To complete a task, the agent must reach the specified destinations in order. 
An independent evaluator is used to verify arrivals by comparing the player's position
with the annotated waypoint coordinates. It samples the client player position
at 1~Hz throughout execution, including during model inference.
The agent is considered to have reached the next required waypoint $g_i$ when its sampled position $p\in\mathbb{R}^3$ satisfies
\[\lVert p-g_i\rVert_2 < 3.5
\quad \text{and} \quad  |p^y-g_i^y|\leq 1.5\]
All distances are measured in Minecraft blocks. The vertical constraint reduces
ambiguity between nearby floors. Only the next required waypoint is eligible
for credit; out-of-order visits do not advance task progress.
An episode can terminate early with failure under three conditions:
  (1) \emph{Death}: the player's health reaches zero after taking damage,
  for example from a fall;
  (2) \emph{Leaving the game}: the player exits or disconnects from the
  game world;
  (3) \emph{Out of bounds}: the player falls beyond the map boundary and
  cannot return to continue the task. On maps with a configured height
  boundary, this is detected when the player remains at or below that
  height for 15 consecutive seconds.
After visiting all required destinations, the agent must submit an accepted
\texttt{claim\_done} to complete the task successfully.

\subsection{Agent Framework}\label{sec:agent-loop} 
\begin{figure}[t]
\centering
\includegraphics[width=\linewidth]{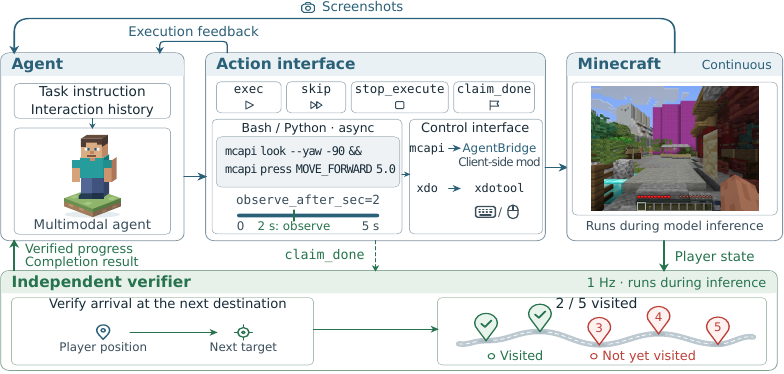}
\caption{
Agent interaction and independent verification in MINE ODYSSEY. }  \vspace{-\baselineskip} 
\label{fig:agent_loop_overview}
\end{figure}

Figure~\ref{fig:agent_loop_overview} shows how the agent interacts with the
Minecraft client through the action interface and receives feedback from the
independent verifier.
 \paragraph{Action Space}
To enable the agent to control a player and interact with the Minecraft client, we develop a programmable control interface that combines API-based player control with keyboard and mouse input.
Specifically, we implement \emph{AgentBridge}, a client-side mod that uses native Minecraft APIs to support player actions such as movement, camera rotation, and item interaction. A  command-line client, \emph{mcapi}, sends commands to AgentBridge and returns execution results to the agent. 
To complement these actions with GUI interactions, we provide \emph{xdo}, a command-line wrapper around \texttt{xdotool} that enables the agent to operate the client’s graphical interface through keyboard and mouse.
The agent uses these tools through Bash, either executing individual operations or composing multiple operations within a single Bash tool call with shell commands or Python scripts. For example, the agent may run \texttt{mcapi look --yaw -90 \&\& mcapi press MOVE\_FORWARD 2.0} to turn the camera and then move forward for two seconds. Scripts can include calculations, loops, and conditional logic, allowing the agent to execute complex action sequences.

\paragraph{Observation Space}
The agent receives screenshots of the Minecraft client as, showing either the player’s first-person view or an open interface, such as the map or inventory. The in-game map provides a top-down view of the explored environment and supports panning, zooming or searching to inspect different areas.  In addition to screenshots, the agent receives textual feedback from command execution, including returned data, execution status, and error messages.
\paragraph{Agent Loop}

Given these observation and control interfaces, the agent interacts with the environment through an observation--action loop. The game simulation runs asynchronously with this loop: the environment evolves continuously without waiting for the agent to finish reasoning or issue its next action. At each decision step, the agent selects one of four action types based on the latest observations, the task instruction, and its interaction history: 
(1)\emph{exec}: execute a Bash command, interrupting any command still running. It can run game-control tools such as \texttt{mcapi} and \texttt{xdo}, execute Python scripts, or perform computations and file operations. Invalid Bash commands return textual errors, while \texttt{mcapi state} queries return textual results; both are returned immediately without waiting for the observation interval.
(2)\emph{skip}: wait for further observations while allowing the current command to continue.
(3)\emph{stop\_execute}: terminate the current command and immediately return a screenshot.
(4)\emph{claim\_done}: indicate that the agent believes it has visited all required waypoints.

Since the environment evolves independently of the agent’s decision loop, conditions may change during command execution, especially when a command takes a long time to complete. The agent may therefore need to observe these changes and intervene before the command finishes. So we provide an optional \texttt{observe\_after\_sec} parameter for \texttt{exec} and \texttt{skip}, which specifies the requested delay in seconds before the next observation. It allows the agent to assess intermediate progress and decide whether to let the current command continue, stop it, or replace it.  For example, the agent may issue
\texttt{exec(content="mcapi press MOVE\_FORWARD 5.0", observe\_after\_sec=2.0)},
  requesting five seconds of forward movement and another observation
  after two seconds. If the path remains clear, it may select \texttt{skip} to wait for another observation while the movement continues. If an obstacle appears, it may select \texttt{stop\_execute} to terminate the command or issue a new \texttt{exec} action to interrupt it and adjust its movement.
\paragraph{Preventing Hacking and Sandbox}
 To preserve the integrity of the evaluation, we restrict access to information and game features that could allow the agent to bypass the intended task requirements. Specifically, we add safeguards to AgentBridge to prevent coordinate disclosure through waypoint editors, settings pages, and chat sharing. The visual world map remains available, while waypoint teleportation and Baritone-based automatic navigation are disabled during evaluation.

In addition, to prevent direct access to hidden task data through Bash, 
{we execute agent-issued commands in a Bubblewrap sandbox  within a containerized evaluation environment managed by Podman or Docker.
  The sandbox uses Linux namespaces and restricted filesystem mounts to expose the required tools and a writable agent workspace while excluding world saves, waypoint files,
  project files, and evaluator records. A separate network namespace disables external network access, while a Unix-domain socket relay permits access to approved
  AgentBridge APIs.}
Such sandbox isolation supports parallel evaluation. Each task runs in its own sandbox with a separate copy, client directory, and agent workspace, preventing interference from others. To improve evaluation efficiency, we also support GPU-accelerated rendering within the containers to offload graphics processing from CPU.
 
\vspace{-1.0pt}
\section{Experiments}
\label{sec:experiments}
\subsection{Setup}
\paragraph{Models and configuration.}
We evaluate leading proprietary models, including GPT-6 Astra~\citep{openai2026gpt6astra},
GPT-5.6 Sol~\citep{openai2026gpt56sol}, Claude Opus 5~\citep{anthropic2026claudeopus5}, Claude Opus 5.5~\citep{anthropic2026claudeopus55},
Gemini 3.8 Flash~\citep{deepmind2026gemini38flash} and Grok 4.6~\citep{xai2026grok46}, alongside open-weight models
DeepSeek-V4.1-Flash~\citep{deepseek2026v41flash}
and GLM-5.3-Flash~\citep{zai2026glm53flash}. All models use native tool calls through
the same action interface. Each task is limited to
500 agent steps and six hours of wall-clock time, with at most
three \texttt{claim\_done} calls.
{ To manage context growth,  we automatically trigger summarization when any of the following three conditions is met: the current context window reaches 100 assistant turns, the latest model call
  reports more than 200K total tokens, or the active context has more than 100 images.  The agent is prompted to summarize its goals, progress, relevant environment state,
  failed attempts, and next steps. The summary replaces the previous interaction history, while the system instructions and original task are retained. The agent then
   resumes the same episode from the current game state using the summary and newly received observations.}
  \paragraph{Tasks and metrics.}
  We evaluate each model on 180 tasks and report \textbf{Success Rate}
  (SR), the fraction of tasks where the agent visits all required stops
  in order and submits an accepted \texttt{claim\_done}.
  We also report \textbf{Checkpoint Coverage} (CC), the average fraction
  of ordered intermediate stops
  to measure partial progress.
  To measure movement and route efficiency, we report \textbf{Path Length}
  (PL) and \textbf{Success weighted by Path Length} (SPL).
  PL averages the full-episode 3D distance, computed from consecutive
  position samples at approximately 1 Hz, over all 180 tasks including failures. 
  {SPL measures efficiency by rewarding successful completion along shorter routes.
  It averages $S_i L_i/\max(L_i,P_i)$ over the same tasks, where $S_i$
  indicates success, $P_i$ is the recorded path length, and $L_i$ is the
  reference path length through all required stops in order}.  Details of reference path computation are provided in
  Appendix~\ref{app:spl-references}.
  Average turns is the average number of agent turns following the agent loop in Section~\ref{sec:agent-loop} and exclude
  automatic history-summary responses.
\subsection{Main results}
Table~\ref{tab:main-results} shows that GPT-6 Astra leads at 85.6\% SR,
followed by Claude Opus 5.5 at 73.9\%. Opus 5.5 exceeds Opus 5 max
by 23.3 percentage points, placing it substantially closer to Astra
than the remaining models. Five of the eight models complete fewer
than half of the tasks; the strongest evaluated open-weight model,
DeepSeek-V4.1-Flash, reaches 23.9\%.
All configurations have lower SR indoors. Astra's lead over Opus 5.5
widens from 8.8 percentage points outdoors to
18.2 points indoors. Opus 5.5 nevertheless
substantially improves on Opus 5 max in both environments, especially
indoors, where SR rises from 32.7\% to 63.6\%. These are comparisons
between task subsets with different layouts and route composition;
they do not isolate the effect of being indoors.
Checkpoint coverage and SPL reveal smaller
gaps between the two leading models. Opus 5.5 reaches 86.1\% CC and
50.35\% SPL, only 2.9 and 3.35 percentage points below Astra,
respectively, despite an 11.7-point SR gap. High intermediate-stop
coverage therefore need not translate into full task completion.
Their mean travel distances are also similar, but Opus 5.5 uses about $2.4\times$ as many decisions. These full-cohort averages include failures and
do not compare routes on a shared set of successful tasks. Gemini
provides a further contrast: it has the shortest mean travel distance but the most decisions and only 26.7\% SR.
Short travel alone therefore does not imply efficient completion.
Within Opus 5, increasing effort from low to max raises SR from
35.0\% to 50.6\% and SPL from 7.77\% to 24.29\%, while mean travel
falls from 4454.2 to 2020.2 blocks and decisions rise from 218.8 to
242.8. Higher effort thus accompanies more completed tasks and less
travel, with a modest increase in decisions.

\begin{table}[t]
    \centering\small
    \setlength{\tabcolsep}{2.5pt}
    \caption{Main results on 180 tasks, with success rates broken down by
    125 outdoor and 55 indoor tasks. SR, CC, and SPL are percentages; PL is
    the mean sampled 3D distance over all tasks.}
    \label{tab:main-results}
    \begin{tabular*}{\linewidth}{@{\extracolsep{\fill}}lrrrrrrr}
        \toprule
        Model & \shortstack{SR (All)$\uparrow$} & \shortstack{SR (Outdoor)$\uparrow$} & \shortstack{SR (Indoor)$\uparrow$} & CC $\uparrow$ & SPL $\uparrow$ & Turns & {PL} \\
        \midrule
        \noalign{\vskip 2pt\hbox to \linewidth{\hfil\strut\textit{Proprietary models}\hfil}\vskip 2pt}
        GPT-6 Astra (medium) & \textbf{85.56} & \textbf{87.20} & \textbf{81.82} & \textbf{89.04} & \textbf{53.70} & 105.8 & 1033.3 \\
        Claude Opus 5.5 (max) & 73.89 & 78.40 & 63.64 & 86.11 & 50.35 & 250.6 & 1051.7 \\
        Claude Opus 5 (max) & 50.56 & 58.40 & 32.73 & 67.53 & 24.29 & 242.8 & 2020.2 \\
        Claude Opus 5 (high) & 44.44 & 51.20 & 29.09 & 59.99 & 17.09 & 231.4 & 2921.5 \\
        Claude Opus 5 (medium) & 39.44 & 47.20 & 21.82 & 57.48 & 12.95 & 233.8 & 3311.3 \\
        Claude Opus 5 (low) & 35.00 & 39.20 & 25.45 & 48.00 & 7.77 & 218.8 & 4454.2 \\
        GPT-5.6 Sol (max) & 43.89 & 51.20 & 27.27 & 61.36 & 22.87 & 306.0 & 1252.4 \\
        Gemini 3.8 Flash (high) & 26.67 & 28.00 & 23.64 & 56.32 & 16.91 & 410.0 & 620.1 \\
        Grok 4.6 (xhigh) & 17.22 & 20.80 & 9.09 & 40.36 & 7.84 & 402.1 & 1088.4 \\
        \midrule
        \noalign{\vskip 2pt\hbox to \linewidth{\hfil\strut\textit{Open-weight models}\hfil}\vskip 2pt}
        DeepSeek-V4.1-Flash (max) & 23.89 & 30.40 & 9.09 & 44.37 & 11.62 & 362.5 & 1327.2 \\
        GLM-5.3-Flash (max) & 15.56 & 19.20 & 7.27 & 36.09 & 5.72 & 357.8 & 1435.6 \\
        \bottomrule
    \end{tabular*}     \vspace{-\baselineskip} 
\end{table}

\section{Analysis}
\subsection{Ablations}
\label{subsec:ablations}

\begin{figure}[t]
    \centering
       \includegraphics[width=0.9\linewidth]{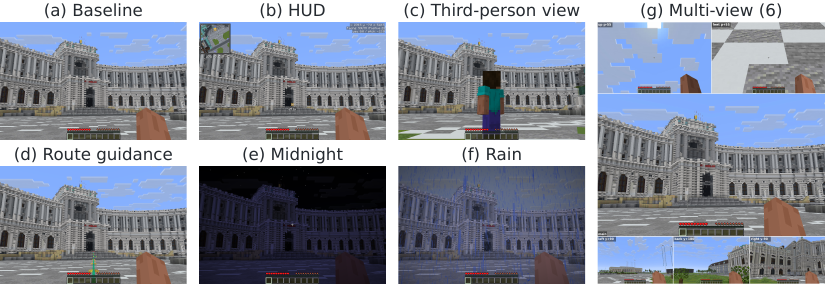}
    \caption{The default view and six ablation conditions are shown outside Hofburg
    Palace.}         
    \vspace{-\baselineskip} 
    \label{fig:ablations}
    
\end{figure}

 We evaluate how spatial information and scene appearance affect
DeepSeek-V4.1-Flash and GLM-5.3-Flash under six conditions. As illustrated in  Figure~\ref{fig:ablations},
{HUD enable BoccHUD (player position and orientation in the upper right) and Xaero's Minimap (local map in the upper left) and remove the default
32-block display-distance limit on first-person waypoint markers.
Multi-view observations combine six camera directions into one image,
while third-person view fixes the camera behind the player.}
{Route guidance display a Baritone-planned path  as a green trail with yellow directional arrows on the ground, as shown in Figure~\ref{fig:ablations} (d),
while leaving movement under agent control.
Midnight replaces the default clear-noon lighting and supplies a torch
that illuminates the surroundings when held.
Rain changes the weather only for outdoor tasks; indoor outcomes remain
unchanged. }

\newpage
\begin{wraptable}{r}{0.64\textwidth}
     \setlength{\abovecaptionskip}{0pt}

    \centering
    \centering
    \caption{Ablation across six conditions: SR, CC, SPL.}
    \label{tab:ablations}
    \footnotesize
    \setlength{\tabcolsep}{2pt}
    \begin{tabular*}{\linewidth}{@{\extracolsep{\fill}}lccc@{}}
        \toprule
        Setting & SR & CC & SPL \\
        \midrule
        \noalign{\vskip 1pt\hbox to \linewidth{\hfil\strut\textit{DeepSeek-V4.1-Flash}\hfil}\vskip 1pt}
        HUD & $30.0\;(+6.1)$ & $52.1\;(+7.8)$ & $12.8\;(+1.2)$ \\
        Multi-view (6) & $26.1\;(+2.2)$ & $48.7\;(+4.3)$ & $11.3\;(-0.3)$ \\
        Third-person view & $28.5\;(+4.5)$ & $50.0\;(+5.3)$ & $13.5\;(+1.8)$ \\
        Route guidance & $21.1\;(-2.8)$ & $46.3\;(+1.9)$ & $9.7\;(-1.9)$ \\
        Midnight & $22.2\;(-1.7)$ & $46.3\;(+2.0)$ & $10.1\;(-1.5)$ \\
        Rain & $20.0\;(-3.9)$ & $42.1\;(-2.3)$ & $9.9\;(-1.7)$ \\
        \midrule
        \noalign{\vskip 1pt\hbox to \linewidth{\hfil\strut\textit{GLM-5.3-Flash}\hfil}\vskip 1pt}
        HUD & $23.0\;(+4.0)$ & $43.8\;(+3.2)$ & $9.6\;(+2.5)$ \\
        Multi-view (6) & $20.6\;(+2.2)$ & $43.3\;(+3.1)$ & $8.6\;(+1.8)$ \\
        Third-person view & $23.0\;(+4.0)$ & $42.8\;(+3.6)$ & $9.9\;(+2.8)$ \\
        Route guidance & $18.3\;(+0.0)$ & $41.8\;(+1.6)$ & $8.0\;(+1.2)$ \\
        Midnight & $16.7\;(-1.7)$ & $39.0\;(-1.2)$ & $7.8\;(+1.0)$ \\
        Rain & $16.1\;(-2.2)$ & $40.3\;(+0.1)$ & $5.7\;(-1.1)$ \\
        \bottomrule
    \end{tabular*} 
\end{wraptable}
All conditions preserve the required destinations. From Table~\ref{tab:ablations}, 
we can find that visual aids improve SR and CC for both models, with HUD raising SR
by 6.1 points for DeepSeek and 4.0 for GLM.
DeepSeek's multi-view SR gain of 2.2 points accompanies a
  0.3 point SPL decrease.  Route guidance slightly raises CC but lowers
  DeepSeek's SR by 2.8 points and leaves GLM's unchanged, indicating
 a gap between
recognizing guidance and effective action.
Finally, midnight
reduces SR by 1.7 points for both models, with torches available to
provide local light. Rain lowers SR by 3.9 points for DeepSeek
and 2.2 for GLM.

\vspace{-1pt}
\subsection{Prolonged local stays.}\label{sec:prolong}
\begin{wrapfigure}{R}{0.56\textwidth}
    \centering
    \includegraphics[width=\linewidth]{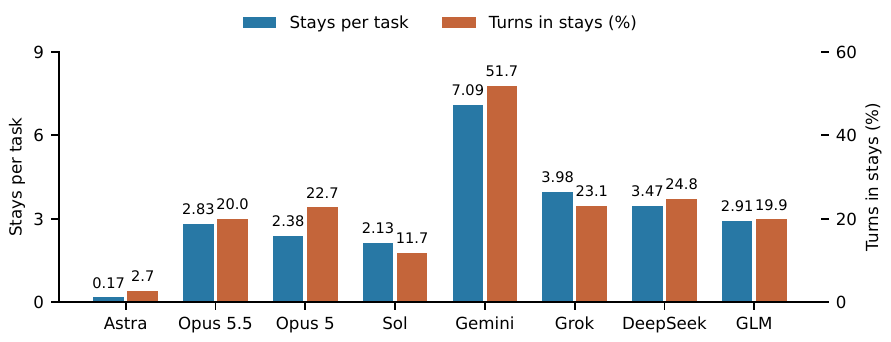}
    \caption{Local stays over at least ten turns within
    one block. Blue: episodes per task (left axis).
    Orange: task-mean share of aligned turns (right axis).}
    \label{fig:stationary-turns}
\end{wrapfigure}
To examine agents' ability to sustain progress, we count
  local stays of at least ten consecutive turns within one block of a
  fixed 3D anchor.
From Figure~\ref{fig:stationary-turns}, we find that Astra rarely remains
  in the same local area for many turns with only 0.17 episodes and 2.7\% of turn, whereas such stays are
  much more frequent among all other models.
 Notably, Gemini spends
  over half of its turns in local stays at 51.7\%, more
  than twice the share of any other model, and averages 7.09 episodes per
  task. During these stays, Gemini often engages in repeated map inspection, menu interactions, and screenshot processing without resuming movement.
\begin{figure}[h]    \centering
  \includegraphics[width=0.8\linewidth]{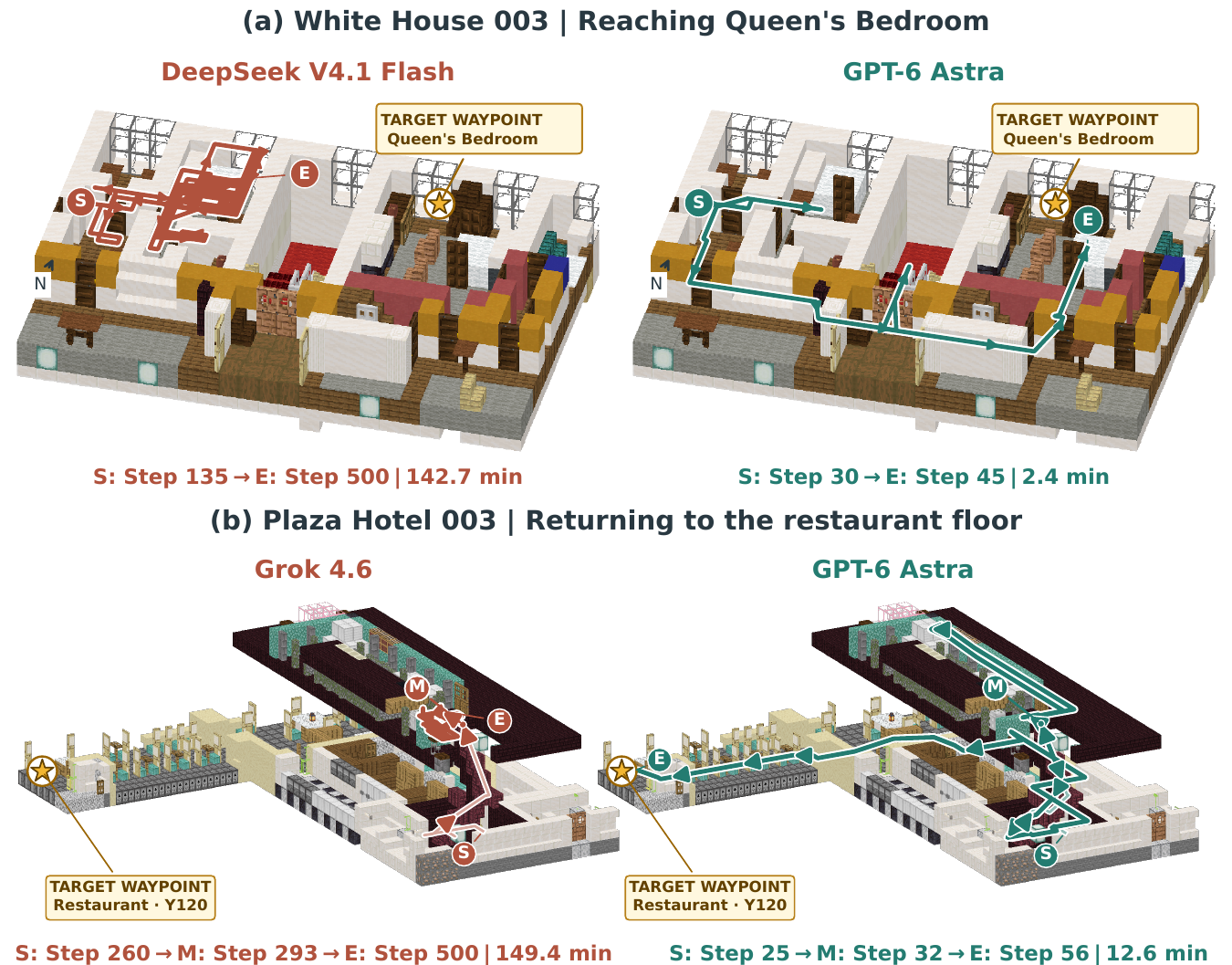}
        \caption
   {(a) White House: Astra takes a detour; DeepSeek revisits rooms.
  (b) Plaza Hotel: Astra returns downstairs; Grok remains upstairs.
  S/E: start/end; M: shared upper landing; stars: targets.
  Times cover the S--E segments.}
 \vspace{-\baselineskip} 
    \label{fig:recovery-comparison}
\end{figure}
\subsection{Case Studies}
\paragraph{Case Study 1}
Figure~\ref{fig:recovery-comparison} shows two pairs of agents
that pass through a shared location but differ in their subsequent progress. 
As shown in Figure~\ref{fig:recovery-comparison}(a), both DeepSeek and Astra enter through the same western
doorway. DeepSeek recognizes the blocked
passage but spent 142.7 minutes repeatedly revisiting the nearby rooms until the step limit.
 Astra instead inspects its
surroundings, returns toward a southern doorway, and follows the hallway
around the wall to Queen's Bedroom, completing the displayed detour in
2.4 minutes.
In Figure~\ref{fig:recovery-comparison}(b),  Astra and Grok both leave the restaurant floor at S
and explore the upper patio at M. Grok remains upstairs for 135.7 minutes before reaching the step limit.  Astra instead retraces its route to the stairs, descends to the restaurant floor, and reaches the target in 12.6 minutes.
In both cases, Astra redirects its search and follows an alternative route to the target, while the unsuccessful models continues exploring within the same local region.
\paragraph{Case Study 2}

\begin{figure}[h]
    \centering
\includegraphics[width=0.85\linewidth]{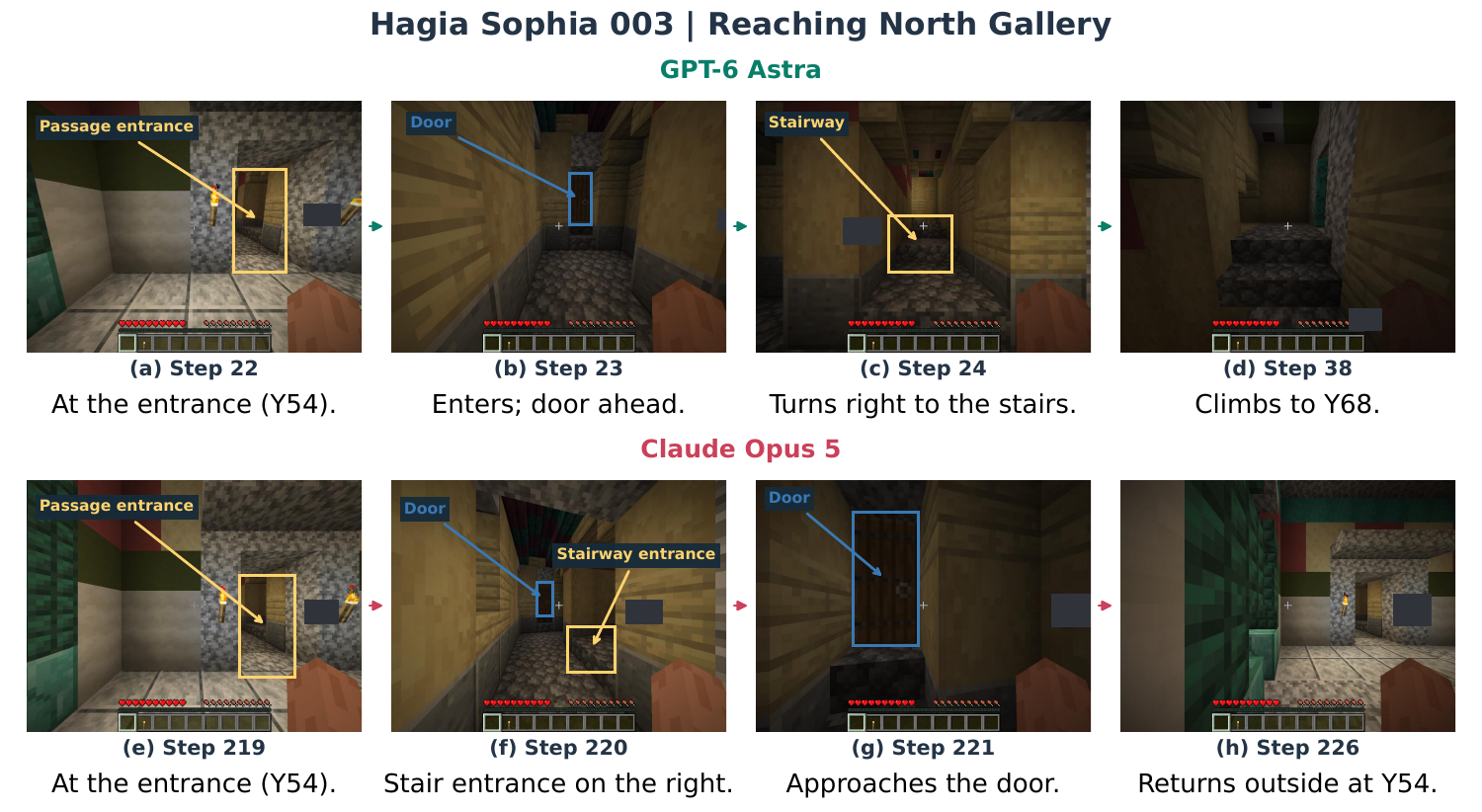}
    \caption{Discovering an upstairs route in Hagia Sophia 003.
  Astra explores the side branch and ascends, while Opus 5 approaches
  the end door and leaves without exploring the branch.} \vspace{-\baselineskip} 
    \label{fig:spatial-reasoning}
\end{figure}
 As illustrated in Figure~\ref{fig:spatial-reasoning}, Astra  and Opus 5 
receive nearly matching views of the same passage entrance. After entering the passage,
they see the same door ahead
(outlined in blue). Astra turns right in place, revealing the stairs and identifying an upward route toward the target. While Opus 5 instead approaches the door and returns outside without exploring the ascending branch, even though the stair opening is visible on its right (outlined in gold). This contrast illustrates how incomplete local exploration can leave an available connection undiscovered, even when its entrance is visible.

\section{Conclusion}

We introduce \benchmark{}, a benchmark for evaluating agentic spatial
intelligence on \numtasks{} spatial tasks spanning
\nummaps{} Minecraft reconstructions of diverse real-world locations.
Our evaluation reveals substantial performance gaps: GPT-6 Astra achieves
85.6\% success, while the second-best model, Claude Opus 5.5, reaches
73.9\% and the strongest evaluated open-weight model, DeepSeek-V4.1-Flash,
reaches 23.9\%, leaving considerable room for improvement.
Through detailed analyses, we identify key challenges in grounding spatial
information, discovering traversable connections, and recovering from
unsuccessful routes. We believe \benchmark{} will support the development
of more capable and reliable agents for sustained interaction in complex
spatial environments.

\bibliography{iclr2027_conference}
\bibliographystyle{iclr2027_conference}
\appendix

\newpage

\section{Related Work}

\subsection{ Spatial Intelligence Evaluation    }
Spatial intelligence evaluation examines how models infer relations,
integrate observations, and use spatial representations. 
Research has expanded from spatial relations in individual images~\citep{du2024embspatial,chen2024spatialvlm} to integration across
views and videos~\citep{yang2024vsibench,yang2025mmsibench,sikarwar2026cosmic},
with finer diagnoses of perspective-taking and rotation~\citep{huang2026spatialdise,zhang2026spinbench}. 
Map understanding and reconstruction further test explicit spatial
representations~\citep{bhat2026mapverse,petersson2025blueprintbench,
andonlabs2026blueprintbench2}. 
Recent interactive benchmarks extend this line to scene refinement~\citep{liu2026spatialact}, active evidence gathering~\citep{hong2026esibench}, and spatial task completion across
simulators~\citep{gao2026spatialworld}. 
Within this line,  we study agentic spatial intelligence in goal-directed navigation, examining how agents align allocentric maps with egocentric observations, select feasible actions under physical constraints, and revise decisions as their actions reveal new evidence.

\subsection {Embodied Navigation Evaluation }
Embodied navigation evaluates whether agents can reach specified goals
  in 3D environments through perception and action.
Instruction-following and object-goal tasks~\citep{batra2020objectnav} evaluate navigation
to individual destinations.
Multi-goal and long-horizon benchmarks~\citep{wani2020multion,khanna2024goat,song2025lhpr,wang2026lmee}
further evaluate whether agents can complete successive navigation stages, adjusting their plans as new observations reveal obstacles, wrong turns, or failed movements.
These navigation tasks are studied in both indoor and outdoor environments.
Indoor benchmarks evaluate navigation through reconstructed rooms and buildings, including low-level movement in continuous environments~\citep{savva2019habitat,krantz2020vlnce}.
For outdoor navigation, street-view environments provide access to real streets but restrict traversal to sampled viewpoints~\citep{chen2019touchdown,mirowski2019streetlearn,yang2024virl}.
City simulators enable continuous navigation through generated urban
scenes~\citep{zhuang2025simworld} and real-city reconstructions such as
UrbanGround's Hong Kong environment, where agents can use maps to navigate ~\citep{ju2026urbanground}.
  \benchmark{} broadens environmental coverage to diverse outdoor
  and indoor reconstructions of real-world places, while opening
  navigation execution to agent-written programs.
  A persistent workspace allows agents to compose navigation
  commands into control routines, process state feedback, and
  retain task information across decisions.

\subsection{Interactive Evaluation of Generalist Agents}
\label{sec:related-agentic-evaluation}
Interactive benchmarks evaluate generalist agents through task completion
in stateful environments~\citep{liu2023agentbench}.
WebArena and OSWorld study realistic web and desktop
workflows~\citep{zhou2024webarena,xie2024osworld}, while OSWorld~2.0
and Toolathlon extend evaluation to sustained execution across longer
workflows~\citep{yuan2026osworld2,li2025toolathlon}.
Game benchmarks examine planning, exploration, and control over
extended play~\citep{paglieri2025balrog,zhang2025videogamebench}.
Beyond task design, GameWorld studies how action interfaces affect
agent performance using state-verifiable task
outcomes~\citep{ouyang2026gameworld}.

Within Minecraft, agent methods investigate reusable skills and memory
for extended execution~\citep{wang2023voyager,wang2023jarvis1,liu2025odyssey}.
Benchmarks assess complementary aspects of this execution:
MCU evaluates composable tasks~\citep{zheng2025mcu}, while MineExplorer
tests whether agents can discover and satisfy unstated prerequisites
through exploration, measuring progress with rule-based
milestones~\citep{ju2026mineexplorer}.
In \benchmark{}, agents operate through a programmable
workspace, where they can process spatial evidence, compose control
routines, and retain intermediate information.
Together, interaction logs, movement trajectories, and evaluator-recorded
visits support analysis of how agents use map information, organize
control, and attempt recovery during the task.

\section{Map Collection and Environment Diversity}
\label{app:map-collection}

Table~\ref{tab:inventory} lists task counts for all \nummaps{} maps in the
selected \numtasks{}-task benchmark.

\begin{table}[htbp]
    \centering
    \caption{Formal task counts in the selected benchmark.}
    \label{tab:inventory}
    \begin{tabular}{lr@{\qquad}lr}
        \toprule
        Map & Tasks & Map & Tasks \\
        \midrule
        Buckingham Palace & 5 & Cape Town / Bo-Kaap & 9 \\
        Chain Bridge & 1 & Copacabana Waterfront & 3 \\
        Dublin Departments District & 7 & Entrup & 6 \\
        Hagia Sophia & 3 & Hofburg & 7 \\
        Innopolis & 9 & Memorial Hall Park & 1 \\
        Miljacka Riverside & 4 & Harbor City Shipyard & 11 \\
        Notre-Dame & 4 & NYC 9/11 Memorials & 6 \\
        Ohrid & 5 & Plaza de Mayo & 8 \\
        Plaza Hotel & 6 & Reichstag & 5 \\
        RMS Queen Mary & 10 & RMS Titanic & 6 \\
        Santa Luc\'ia Hill & 5 & Shun Lee & 4 \\
        SoFi Stadium & 6 & Sviyazhsk & 6 \\
        Torrey Mall & 5 & Ueno Park & 12 \\
        Versailles & 3 & White House & 7 \\
        W\"urzburg & 10 & Z\"urich & 6 \\
        \midrule
        \multicolumn{3}{r}{Total} & 180 \\
        \bottomrule
    \end{tabular}
\end{table}

The collection represents locations in countries and regions including
South Africa, Brazil, Ireland, Japan, Germany, France, T\"urkiye,
the United Kingdom, the United States, and Hong Kong.
Examples range from city districts in Cape Town, Rio de Janeiro, Dublin,
Tokyo, and New York City to Buckingham Palace, Versailles, Hagia Sophia,
the Reichstag, SoFi Stadium, and historic passenger ships.
Outdoor maps cover residential and commercial districts, parks, hillsides,
and waterfronts, with variation in street connectivity, landmark density,
and relief. Indoor maps include palaces, religious buildings, hotels,
stadiums, and passenger ships, varying in room arrangement, corridor
structure, entrances, and vertical circulation. Together, these settings
support tasks from route planning across city districts to finding
building entrances and moving between floors. Stairs, ladders, doors,
and gates provide additional local traversal and interaction contexts,
as illustrated in Figure~\ref{fig:terrain-interactions}. In Figure~\ref{fig:map-overview}, we present in-game world-map views of all
\nummaps{} benchmark environments.
Each map is labeled with its name and independently scaled to show its
full layout while preserving its aspect ratio.

\begin{figure}[t]
    \centering
    \includegraphics[width=\linewidth]{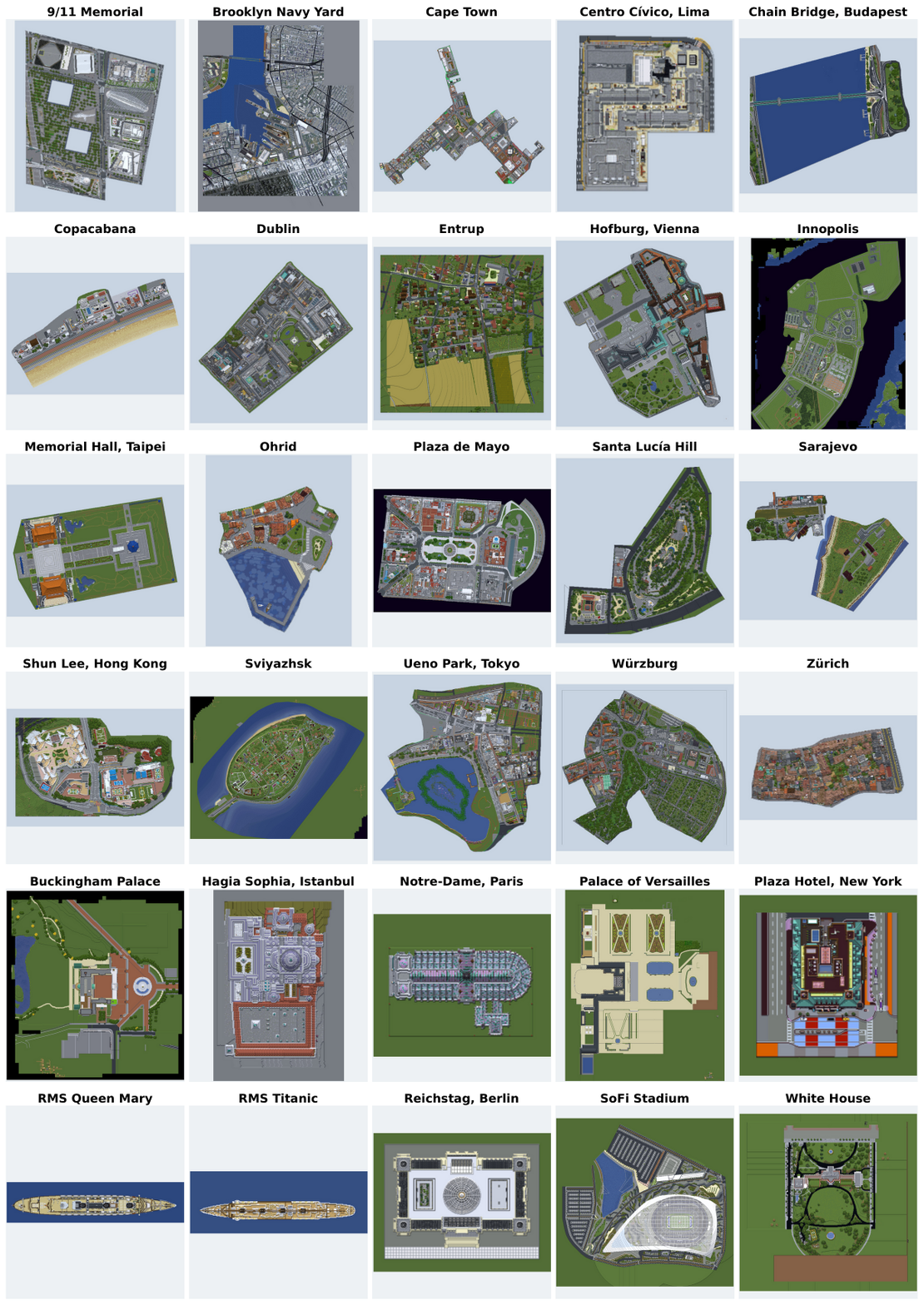}
    \caption{In-game world-map views of all \nummaps{} benchmark environments.
   }
    \label{fig:all-maps}
\end{figure}
\section{Default Settings and Game Configuration}
\label{fig:map-overview}

\subsection{Default settings.}\label{app:settings}
Table~\ref{tab:default-settings} summarizes the settings shared by the
main-comparison runs. Controlled variants modify the designated settings
as described in Section~\ref{subsec:ablations}.

\begin{table}[htbp]
    \centering
    \caption{Default settings for the main comparison. Controlled variants
    override the relevant entries.}
    \label{tab:default-settings}
    \begin{tabular}{ll}
        \toprule
        Factor & Default \\
        \midrule
        Time / weather & Noon (frozen) / clear (frozen) \\
        Perspective & First person \\
        Observation & Event-based, single view \\
        Player scale & 1.0 \\
        Optional HUD / minimap panel & Disabled \\
        Full-screen world map & Available \\
        Navigation hints & Disabled \\
        Target-coordinate access & Locked in client and chat-sharing paths \\
        Action-sandbox network & Public IP access disabled \\
        Game mode / difficulty & Adventure / peaceful \\
        Mobs / initial inventory & Disabled / cleared \\
        Time / decision limits & 21,600 seconds / 500 decisions \\
        \bottomrule
    \end{tabular}
\end{table}
\paragraph{Control and condition-specific configurations.}
Our \emph{AgentBridge} mod (1.0.0) provides player control.
In the midnight condition, \emph{LambDynamicLights} provides illumination
from held torches. The route-guidance condition additionally loads our
\emph{Ground Navigation} mod (0.1.0) and \emph{Baritone} (1.17.0)
to display a planned route while leaving movement under agent control
(Section~\ref{subsec:ablations}).
\subsection{Observation and interaction.}
The default observation is a single first-person view. Observations and
execution feedback are returned through the agent's interaction loop;
the game and verifier continue to run during model inference. Agents can
request an observation delay through \texttt{observe\_after\_sec}, allowing
inspection before a command finishes. State queries and execution errors
can also return textual feedback. Thus, event-based observation does not
mean that the world pauses between decisions.
Adventure mode permits ordinary locomotion and interactions such as
opening doors and climbing ladders, while restricting block modification.
Flight is disabled, and Peaceful difficulty removes hostile combat.
Agents may consult the full-screen world map and nearby waypoint markers,
but cannot access exact target coordinates or evaluator state.
The minimap panel, optional HUD, navigation hints, automated navigation,
and waypoint teleportation are disabled by default. Agent-issued commands
run in an isolated workspace without access to world saves, waypoint
files, or evaluator records; a restricted local relay provides access to
the approved game-control APIs.

\subsection{SPL and reference path construction}
\label{app:spl-references}

Success weighted by Path Length (SPL) measures navigation efficiency by
combining task success with path length relative to a reference:
\begin{equation}
    \mathrm{SPL} = \frac{1}{N}\sum_{i=1}^{N}
    S_i\frac{L_i}{\max(L_i,P_i)},
\end{equation}
where $N=180$ is the number of tasks, $S_i$ indicates task success,
$P_i$ is the agent's full-episode path length, and $L_i$ is the reference
path length through all required stops in order. Failed tasks contribute
zero; successful tasks receive higher scores when their paths are shorter
relative to the reference. Each task's contribution is capped at one.

We use A*, a shortest-path search algorithm~\citep{hart1968formal},
to compute a reference path for each task, shared across all models.
We represent Minecraft movement as a graph, where nodes denote candidate
player positions and edges denote movements allowed by our approximate
movement model. Each edge is weighted by the 3D distance between its
endpoints. Starting from the recorded initial position, A* finds the
minimum-length route through all required arrival regions in order.
The graph uses a 0.5-block horizontal lattice and approximate collision
geometry. It supports walking, steps,
ladders, and bounded falls with a crouched player body; it accounts for
cumulative fall damage on maps where damage is enabled. Ordinary doors
and registered button-operated gates are modeled as operable.
Gap jumps, swimming, dynamic stance changes, and button-operated
transport are outside the current graph.
Both reference and agent path lengths are approximate. Reference lengths
depend on the graph's discretization and simplified movement model,
while agent path lengths are estimated from
position samples collected at approximately 1 Hz, which may miss movement
between samples.

\section{Performance and Inference Cost}
\label{sec:resource-tradeoffs}
\label{app:resource-accounting}

\subsection{Resource Accounting}

Figures~\ref{fig:cost-performance} and~\ref{fig:resource-tradeoffs}
report estimated monetary cost and output-token usage. Means include
all 180 tasks, including failed runs. Turns count model decisions and
exclude automatic history-summary responses.

\subsection{Performance and Resource Use}

\begin{figure}[t]
    \centering
    \includegraphics[width=0.7\linewidth]{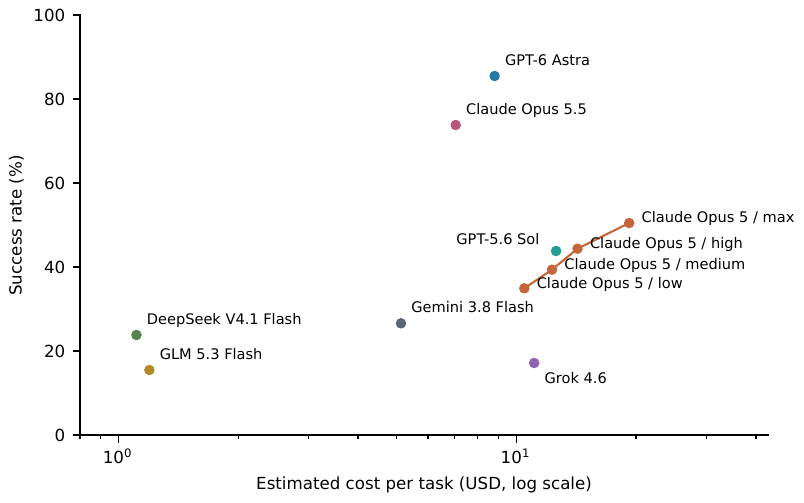}
    \caption{SR versus estimated cost per task, with cost on a log scale.
    Connected points show Claude Opus 5 effort settings. Costs average
    over all 180 tasks, including failures, using the accounting in
    Appendix~\ref{app:resource-accounting}. The horizontal interval indicates
    the estimated cost range for Opus 5.5.}
    \label{fig:cost-performance}
\end{figure}

The cost comparison distinguishes models with similar spending but
different completion rates (Figure~\ref{fig:cost-performance}). At around
\$1 per task, DeepSeek-V4.1-Flash achieves 23.9\% SR for \$1.11, compared
with 15.6\% for GLM-5.3-Flash at \$1.20. DeepSeek therefore achieves
higher completion at slightly lower estimated cost in this comparison.
Moving to Gemini 3.8 Flash raises estimated spending to \$5.13 per task,
about $4.6\times$ as much, while improving SR by 2.8 percentage points.

GPT-6 Astra achieves the highest SR and the lowest reported output-token
usage, reaching 85.6\% SR with 10.3K output tokens per task
(Figure~\ref{fig:resource-tradeoffs}, left). Among the other models,
GPT-5.6 Sol achieves 43.9\% SR with 68.5K output tokens, while DeepSeek
V4.1 Flash generates 618.0K tokens but reaches 23.9\% SR.
Output volume alone therefore does not explain the completion ranking.
This comparison concerns reported output usage, rather than a direct
measurement of inference compute.

\subsection{Reasoning Effort}

\begin{figure}[h]
    \centering
    \includegraphics[width=0.8\linewidth]{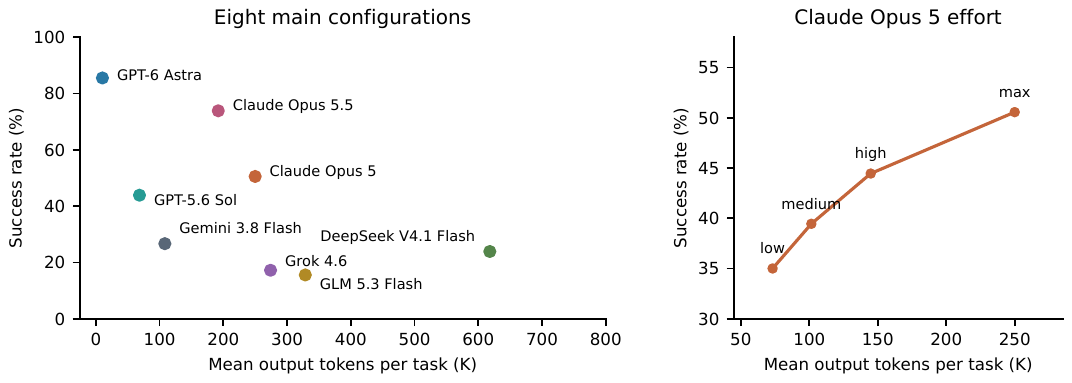}
    \caption{SR versus output-token usage. Left: the eight main model
    configurations. Right: Claude Opus 5 at low, medium, high, and max
    reasoning effort; connected points show increasing effort. Values
    use all 180 tasks, including failures; K denotes thousands.
    Both panels use linear axes, with different SR ranges.}
    \label{fig:resource-tradeoffs}
\end{figure}

Increasing Claude Opus 5's reasoning effort from low to max raises SR
from 35.0\% to 50.6\%, while mean output usage grows from 73.2K to
249.9K tokens per task, approximately $3.4\times$
(Figure~\ref{fig:resource-tradeoffs}, right).
Estimated monetary cost increases from \$10.47 to \$19.21 per task,
yielding a 15.6-point SR gain for an additional \$8.74.
Mean turns increase by only 11.0\%, so output usage grows much faster
than the number of interaction turns.

The additional output required also differs across effort transitions:
low to medium adds 28.3K output tokens per task for a 4.4-point SR gain,
whereas high to max adds 105.1K for a 6.1-point gain
(Figure~\ref{fig:resource-tradeoffs}, right). Higher effort improves completion in this
model, but the latter transition requires more additional output per
percentage point gained. These comparisons average over all tasks;
they do not hold successful-task membership fixed.

\section{Failure Stages}
\label{app:failure-stages}

\begin{figure}[htbp]
    \centering
    \includegraphics[width=0.9\linewidth]{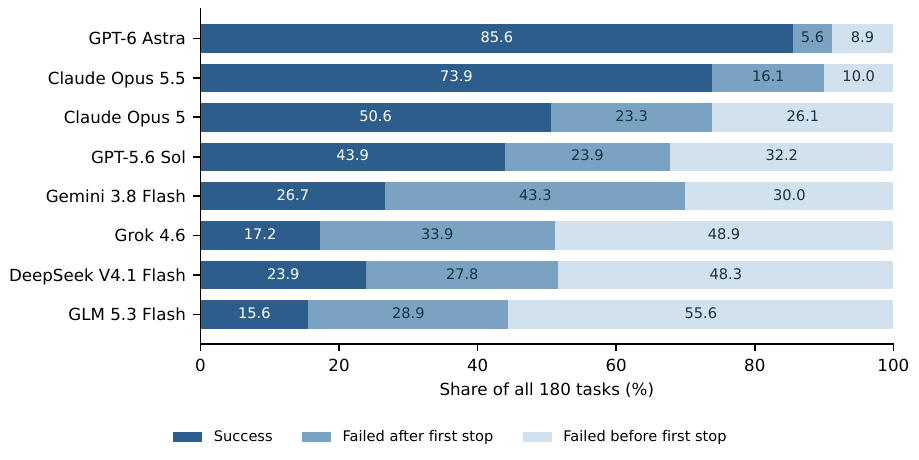}
    \caption{Outcomes over 180 tasks per model. First stop means the
    first intermediate stop, or the destination if none. Deaths count
    as failures even after all stops. All shares use the full 180-task
    denominator. Model configurations match the main comparison.}
    \label{fig:outcome-phases}
\end{figure}

Figure~\ref{fig:outcome-phases} distinguishes failure before the first required stop, failure after reaching it, and successful
completion. The first stop is the first intermediate stop, or the destination for tasks without inter-
mediate stops. Reaching a stop uses the evaluator’s ordered-arrival record; full success additionally
requires an accepted completion claim. Failure after the first stop can therefore include runs that visit
all stops but do not complete successfully.
DeepSeek, Grok, and GLM fail before the first stop on 48.3–55.6\% of all tasks. Gemini reaches it
more often than Sol (70.0\% versus 67.8\%), but fails afterward on 43.3\% of all tasks, compared with
23.9\% for Sol. These outcomes show why initial progress need not translate into full completion.

\section{Interaction Statistics}
\label{app:interaction-statistics}

\begin{figure}[htbp]
    \centering
    \includegraphics[width=\linewidth]{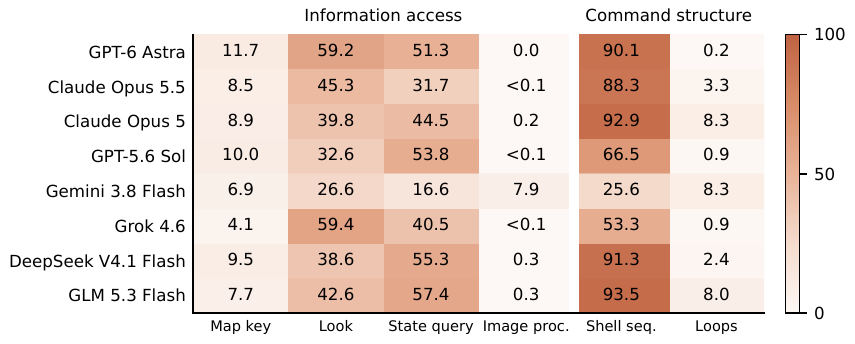}
    \caption{Information access and command structure across eight models.
Values are percentages of execution requests, computed per task and
averaged equally across tasks; categories can overlap.
Map key counts map toggles, Look counts camera-orientation commands,
State query includes queries combined with controls, and Image proc. counts explicit image-processing operations.
Shell sequence counts command sequences or pipelines within a request, and Loops counts requests containing shell or Python loop syntax.}
    \label{fig:interaction-styles}
\end{figure}

\paragraph{Counting and aggregation.}
We classify 400,571 exec requests from the eight configurations,
each evaluated on the same 180 tasks. Each recognized category is counted
at most once per request, even when a request contains repeated calls.
For each category, we divide its count by the task's exec-request count
and then average these percentages equally across tasks. We do
not pool requests across tasks. A request may contribute to multiple
categories, and percentages need not sum to 100\%.
\paragraph{Observed interaction patterns}

As revealed in Figure~\ref{fig:interaction-styles}, Astra, Opus 5, DeepSeek, and GLM combine commands in over 90\% of requests,
 compared with 25.6\% for Gemini. Gemini has the highest share of explicit image processing. It also includes state
queries less frequently. Section~\ref{sec:prolong} also reveals that
Gemini often remains in one place for extended periods, repeatedly checking
maps, operating menus, and processing screenshots without resuming movement.
Opus 5 and Gemini use loops
more often than Astra, whose requests predominantly use command sequences.
Astra and Grok both adjust the viewing direction in about 59\% of requests,
while Astra toggles the map more frequently.
These patterns distinguish how models request information and organize
commands, without directly ranking navigation ability. 
\section{CASE Studies}~\label{sec:case-programmatic-navigation} 

\begin{figure}[h]
    \centering
    \includegraphics[width=\linewidth]{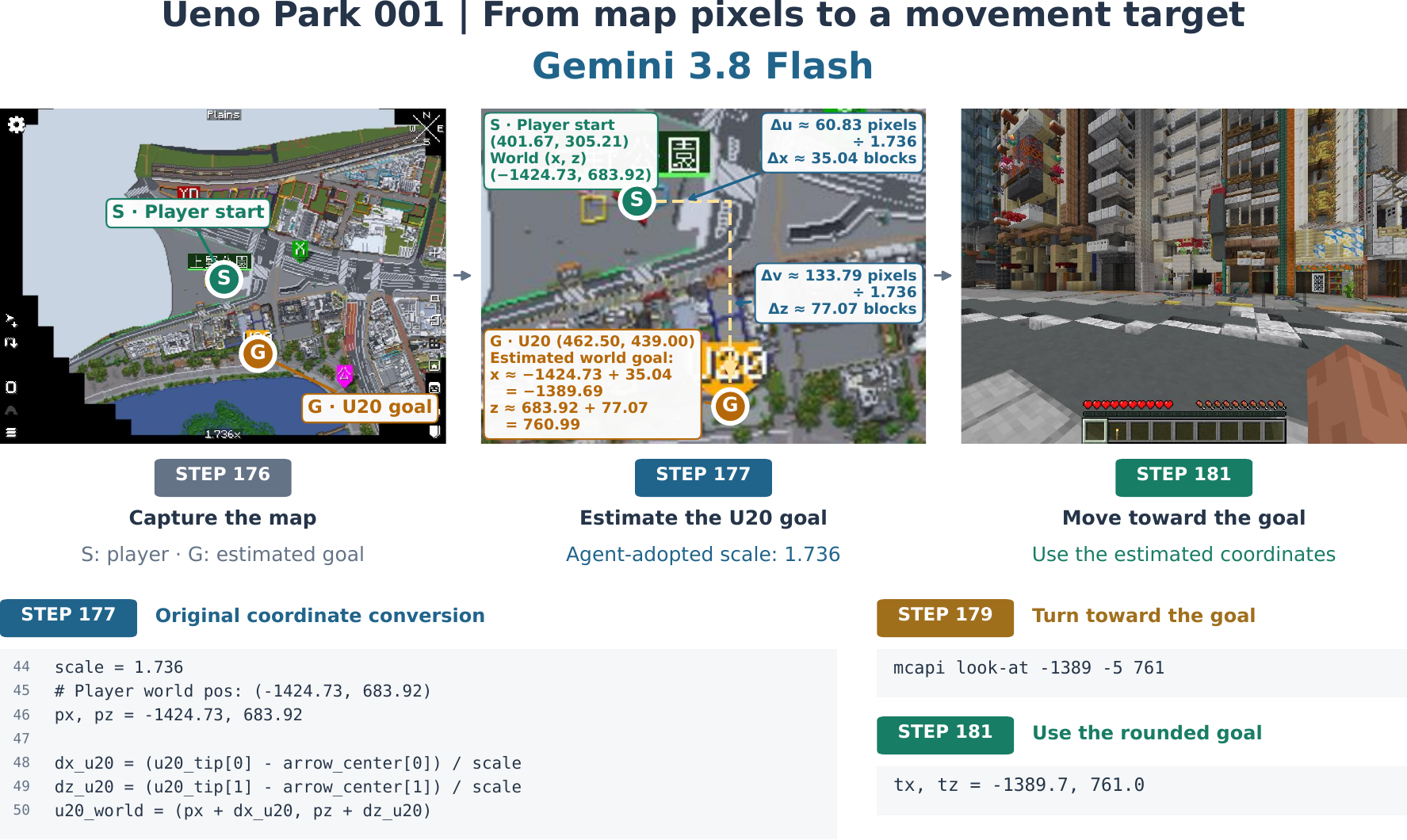}
    \caption{\textbf{From map pixels to a movement target.}
    Step 176 captures the map; Step 177 computes the estimated U20
    coordinates; Steps 179 and 181 use the estimate for orientation and
    movement. S marks the player position and G the estimated goal;
    dashed lines indicate pixel offsets, not an executed route. The
    displayed map provides context because the measured PNG is
    unavailable. The adopted scale is uncalibrated, and execution does
    not establish arrival.}
    \label{fig:programmatic-map}
    \label{fig:map-grounding-case}
\end{figure}
\paragraph{Image processing can construct an actionable spatial target.}
Gemini turns map observations into an actionable spatial target through
a short Python program. It converts player-to-marker pixel offsets into
estimated world coordinates, then uses the result to orient and initiate
movement toward U20 (Figure~\ref{fig:programmatic-map}). This case
illustrates how programming connects visual map grounding to action.

\begin{figure}[t]
    \centering
    \includegraphics[width=0.9\linewidth]{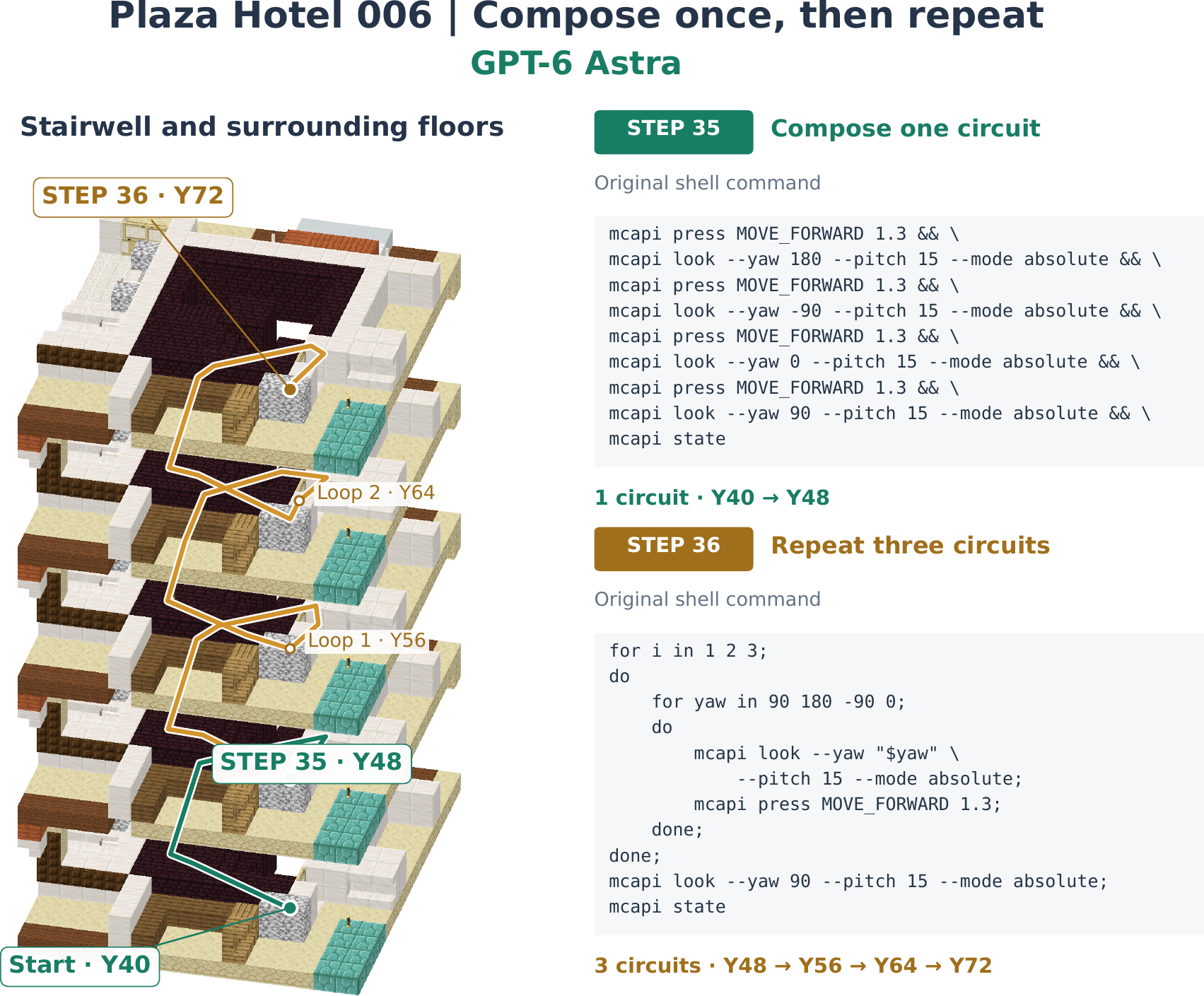}
    \caption{From composition to reuse in Plaza Hotel, Steps 35--36.
    Step 35 executes one circuit from Y40 to Y48; Step 36 executes all
    three circuits from Y48 through Y56 and Y64 to Y72. The cutaway uses
    the retained world's native textures, with selected walls and room
    volumes removed. Command-window position samples and matched state
    endpoints are projected at their original heights over the scene
    for visibility. Matching colors link each step
    in the enlarged scene to its height transition and complete shell
    command. The original observation screenshots are omitted. Only whitespace
    is reformatted in the code. Later steps are outside the displayed
    window; the cutaway is a rendering, not an agent screenshot.}
    \label{fig:programmatic-stairs}
\end{figure}

\paragraph{Stair programming.}
As shown in Figure~\ref{fig:programmatic-stairs},  after testing individual
stair movements, Astra combines four 1.3-second forward movements and
turns into a single shell sequence. At Step 35, this sequence completes
one circuit around the stairwell, raising the agent from Y40 to Y48.
At Step 36, Astra uses a loop to repeat the four-heading movement pattern
three times within one tool request, ascending through Y56 and Y64 to Y72.
The loop reuses the established stair-climbing pattern across successive
levels without requiring a new agent turn for each movement.

Together, these cases show how agents use programming to turn visual
map information into explicit coordinates and individual movement
commands into reusable action sequences.

\end{document}